%% file: arxiv.tex
\documentclass[10pt,twocolumn,letterpaper]{article}

\usepackage{cvpr}
\usepackage{times}
\usepackage{epsfig}
\usepackage{graphicx}
\usepackage{amsmath}
\usepackage{amssymb}

\usepackage{xfrac}
\usepackage{subcaption}
\usepackage[export]{adjustbox}
\usepackage[table]{xcolor}
\usepackage{color}
\usepackage{rotating}
\usepackage{array}

\definecolor{myblue}{rgb}{.1,0.3,0.9}
\definecolor{rowblue}{RGB}{220,230,240}
\usepackage{booktabs}
\usepackage{tablefootnote}
\usepackage{colortbl}
\usepackage{tabularx}
\usepackage{multirow}

\usepackage[pagebackref=true,breaklinks=true,letterpaper=true,colorlinks,bookmarks=false]{hyperref}

\cvprfinalcopy 

\def\cvprPaperID{3310} 

\ifcvprfinal\fi
\begin{document}

\title{Steady-state Non-Line-of-Sight Imaging}

\author{
	Wenzheng Chen$^{1,2}$\thanks{The majority of this work was done while interning at Algolux.}\hspace{0.1in}
	Simon Daneau$^{1,3}$\hspace{0.1in}
	Fahim Mannan$^{1}$\hspace{0.1in}
    Felix Heide$^{1,4}$\hspace{0.1in} \vspace{5pt}
		\\ 
		\textsuperscript{1}Algolux\hspace{0.1in}
	\textsuperscript{2}University of Toronto\hspace{0.1in}
	\textsuperscript{3}Universit\'e de Montr\'eal\hspace{0.1in}
	\textsuperscript{4}Princeton University\hspace{0.1in}
}

\maketitle
\input{definition}

\begin{abstract}
Conventional intensity cameras recover objects in the direct line-of-sight of the camera, whereas occluded scene parts are considered lost in this process. Non-line-of-sight imaging (NLOS) aims at recovering these occluded objects by analyzing their indirect reflections on visible scene surfaces. Existing NLOS methods temporally probe the indirect light transport to unmix light paths based on their travel time, which mandates specialized instrumentation that suffers from low photon efficiency, high cost, and mechanical scanning. We depart from temporal probing and demonstrate steady-state NLOS imaging using conventional intensity sensors and continuous illumination. Instead of assuming perfectly isotropic scattering, the proposed method exploits directionality in the hidden surface reflectance, resulting in (small) spatial variation of their indirect reflections for varying illumination. To tackle the shape-dependence of these variations, we propose a trainable architecture which learns to map diffuse indirect reflections to scene reflectance using only synthetic training data. Relying on consumer color image sensors, with high fill factor, high quantum efficiency and low read-out noise, we demonstrate high-fidelity color NLOS imaging for scene configurations tackled before with picosecond time resolution.
\end{abstract}

\section{Introduction}
\input{introduction}
\section{Related Work}
\input{related_work}

\section{Image Formation Model}
\input{image_formation}

\section{Inverse Indirect Transport for Planar Scenes }
\label{sec:analytic}
\input{planar}

\section{Learning Inverse Indirect Illumination}
\input{method}

\section{Training Datasets}

\input{dataset}

\section{Evaluation}\label{sec:eval}
\input{result}

\section{Conclusion}
\input{discussion}

\section{Acknowledgements}
The authors thank Colin Brosseau for many fruitful discussions and assisting with the experiments. We thank the Vector Institute for supporting Wenzheng Chen.

\clearpage
{\small
\bibliographystyle{ieee}
\bibliography{bib}
}

\end{document}

%% file: definition.tex

\definecolor{Gray}{rgb}{0.5,0.5,0.5}
\definecolor{darkblue}{rgb}{0,0,0.7}
\definecolor{orange}{rgb}{1,.5,0} 
\definecolor{red}{rgb}{1,0,0} 

\newcommand{\heading}[1]{\noindent\textbf{#1}}
\newcommand{\note}[1]{{\em{\textcolor{orange}{#1}}}}
\newcommand{\todo}[1]{{\textcolor{red}{\bf{TODO: #1}}}}
\newcommand{\comments}[1]{{\em{\textcolor{orange}{#1}}}}
\newcommand{\changed}[1]{#1}
\newcommand{\place}[1]{ \begin{itemize}\item\textcolor{darkblue}{#1}\end{itemize}}
\newcommand{\de}{\mathrm{d}}

\newcommand{\Radiance}{\mathit{L}}             
\newcommand{\Refl}{\mathit{\rho}}             
\newcommand{\Geom}{\mathit{G}}             
\newcommand{\geom}{g}             

\newcommand{\volume}{V}
\newcommand{\normal}{\Vect{n}}
\newcommand{\planepos}{\Vect{v}}
\newcommand{\laserpos}{\Vect{l}}               
\newcommand{\wallpos}{\Vect{w}}               
\newcommand{\voxelpos}{\Vect{x}}      
\newcommand{\pixelpos}{\Vect{c}}               
\newcommand{\pathlength}{p}
\newcommand{\posvariable}{\Vect{y}}
\newcommand{\planepoint}{\Vect{p}}
\newcommand{\reflpoint}{\Vect{c}}
\newcommand{\convmatrix}{\mathbf{K}}

\newcommand{\normlzd}[1]{{#1}^{\textrm{aligned}}}

\newcommand{\ttime}{\tau}               
\newcommand{\x}{\Vect{x}}               
\newcommand{\z}{z}               

\newcommand{\npixels}{n}               
\newcommand{\ntime}{t}               

\newcommand{\illfunc}     {g}
\newcommand{\pathfunc}     {s}
\newcommand{\camfunc}     {f}

\newcommand{\pointmult}{\odot} 

\newcommand{\irradiance}{E}
\newcommand{\exposure}{b}
\newcommand{\pmdfunc}{f}                
\newcommand{\lightfunc}{g}              
\newcommand{\period}{T}                 
\newcommand{\freqm}{\omega}                
\newcommand{\illphase}{\rho}             
\newcommand{\sensphase}{\psi}             
\newcommand{\pmdphase}{\phi}            
\newcommand{\omphi}{{\omega,\phi}}      
\newcommand{\numperiod}{N}              
\newcommand{\att}{\alpha}               
\newcommand{\pathspace}{{\mathcal{P}}}  

\newcommand{\atan}{\operatorname{atan}}

\newcommand{\Fourier}{\mathfrak{{F}}}         
\newcommand{\conv}     {\otimes}
\newcommand{\corr}     {\star}
\newcommand{\Mat}[1]    {{\ensuremath{\mathbf{\uppercase{#1}}}}} 
\newcommand{\Vect}[1]   {{\ensuremath{\mathbf{\lowercase{#1}}}}} 
\newcommand{\Id}				{\mathbb{I}} 
\newcommand{\Diag}[1] 	{\operatorname{diag}\left({ #1 }\right)} 
\newcommand{\Opt}[1] 	  {{#1}_{\text{opt}}} 
\newcommand{\CC}[1]			{{#1}^{*}} 
\newcommand{\Op}[1]     {\Mat{#1}} 
\newcommand{\minimize}[1] {\underset{{#1}}{\operatorname{argmin}} \: \: } 
\newcommand{\maximize}[1] {\underset{{#1}}{\operatorname{argmax}} \: \: } 
\newcommand{\grad}      {\nabla}
\newcommand{\cross}      {\times}

\newcommand{\Basis}{\Mat{H}}         		
\newcommand{\Corr}{\Mat{C}}             
\newcommand{\correlem}{\bold{c}}             
\newcommand{\meas}{\Vect{b}}            
\newcommand{\Meas}{\Mat{B}}            
\newcommand{\MeasNormalized}{\Mat{B}^{\textrm{new}}}            
\newcommand{\Img}{H}                    
\newcommand{\img}{\Vect{h}}             
\newcommand{\latentresponse}{\alpha}

%% file: introduction.tex
\begin{figure}[t!]
\vspace{-2pt}
    \centering
		\includegraphics[width=0.95\linewidth]{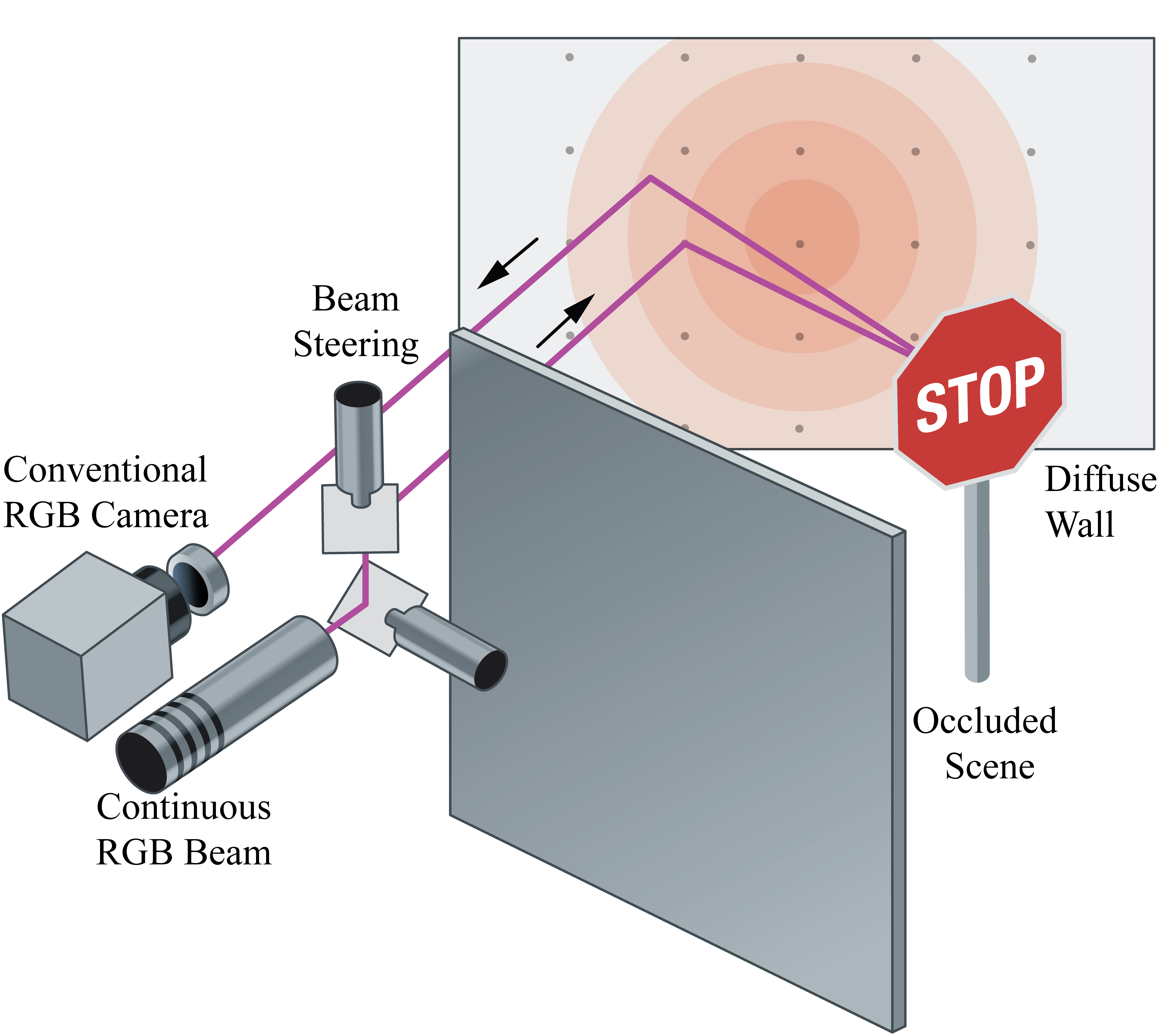}
		\vspace{-6pt}
    \caption{\small We demonstrate that it is possible to image occluded objects outside the direct line-of-sight using continuous illumination and conventional cameras, without temporal sampling. We sparsely scan a diffuse wall with a beam of white light and reconstruct ``hidden'' objects only from spatial variations in steady-state indirect reflections.}
    \label{fig:Teaser}
    \vspace{-10pt}
\end{figure}

Recovering objects from conventional monocular imagery represents a central challenge in computer vision, with a large body of work on sensing techniques using controlled illumination with spatial~\cite{scharstein2003high,o2012primal} or temporal coding~\cite{lange00tof, kadambi2013sparsedeconvolution,heide2013low,otoole2014temporal}, multi-view reconstruction methods~\cite{hartley2003multiple}, sensing via coded optics~\cite{raskar2007computational}, and recently learned reconstruction methods using single-view monocular images~\cite{saxena2006learning,eigen2014depth,godard2017unsupervised}. While these sensing methods drive applications across domains, including autonomous vehicles, robotics, augmented reality, and dataset acquisition for scene understanding~\cite{song2015sun}, they only recover objects in the direct line-of-sight of the camera. This is because objects outside the line-of-sight only contribute to a measurement through indirect reflections via visible diffuse object surfaces. These reflections are extremely weak due to the multiple scattering, and they lose (most) angular information on the diffuse scene surface (as opposed to a mirror surface in the scene). NLOS imaging aims at recovering objects outside a camera's line-of-sight from these indirect light transport components.

To tackle the lack of angular resolution, a number of NLOS approaches have been described that temporally probe the light-transport in the scene, thereby unmixing light path contributions by their optical path length~\cite{Abramson:78:LIF,Kirmani:2009,Naik:2011,Pandharkar:2011} and effectively trading angular with temporal resolution. To acquire temporally resolved images of light transport, existing methods either directly sample the temporal impulse response of the scene by recording the temporal echoes of laser pulses~\cite{Velten:2012recovering,Pandharkar:2011,Gupta:12,buttafava2015non,tsai2017geometry,arellano2017fast,o2018confocal}, or they use amplitude-coded illumination and time-of-flight sensors~\cite{heide2014diffuse,kadambi2016occluded,kadambi2013coded}. While amplitude coding approaches suffer from low temporal resolution due to sensor demodulation bandwidth limitations~\cite{lange00tof} and the corresponding ill-posed inverse problem~\cite{heide2013low}, direct probing methods achieve high temporal resolution already in the acquisition phase, but in turn require ultra-short pulsed laser illumination and detectors with $<$ 10~ps temporal resolution for macroscopic scenes. This mandates instrumentation with high temporal resolution, that suffers from severe practical limitations including low photon efficiency, large measurement volumes, high-resolution timing electronics, excessive cost and monochromatic acquisition. Early streak-camera setups~\cite{Velten:2012recovering} hence require hours of acquisition time, and, while emerging single photon avalance diode (SPAD) detectors~\cite{buttafava2015non,o2018confocal} are sensitive to individual photons, they are in fact photon-inefficient (diffuse experiments in~\cite{o2018confocal}) due to very low fill factors and pileup distortions at higher pulse power. To overcome this issue without excessive integration times, recent approaches~\cite{o2018confocal,heide2017robust} restrict the scene to retro-reflective material surfaces, which eliminates quadratic falloff from these surfaces, but effectively also constrains practical use to a single object class.

In this work, we demonstrate that it is possible to image objects outside of the direct line-of-sight using conventional intensity sensors and continuous illumination, without temporal coding. In contrast to previous methods, that assume perfectly isotropic reflectance, the proposed method exploits directionality of the hidden object's reflectance, resulting in spatial variation of the indirect reflections for varying illumination. To handle the shape-dependence of these variations, we learn a deep model trained using a training corpus of simulated indirect renderings. By relying on consumer color image sensors, with high fill factor, high quantum efficiency and low read-out noise, we demonstrate full-color NLOS imaging at fast imaging rates and in setup scenarios identical to those tackled by recent pulsed systems with picosecond resolution.

Specifically, we make the following contributions:
\begin{itemize}
	\setlength\itemsep{.2em}
	\item We formulate an image formation model for steady-state NLOS imaging and an efficient implementation without ray-tracing. Based on this model, we derive an optimization method for the special case of planar scenes with known reflectance.
	\item We propose a learnable architecture for steady-state NLOS imaging for representative object classes.
	\item We validate the proposed method in simulation, and experimentally using setup and scene specifications identical to the ones used in previous time-resolved methods. We demonstrate that the method generalizes across objects with different reflectance and shapes.
	\item We introduce a synthetic training set for steady-state NLOS imaging. The dataset and models will be published for full reproducibility.
\end{itemize}

%% file: related_work.tex
\noindent\textbf{Transient Imaging}
Kirmani et al.~\cite{Kirmani:2009} first proposed the concept of recovering ``hidden'' objects outside a camera's direct line-of-sight using temporally resolved light transport measurements in which short pulses of light are captured ``in flight'' before the global transport reaches a steady state. These transient measurements are the temporal impulse response of light transport in the scene. Abramson~\cite{Abramson:78:LIF} first demonstrated a holographic capture system for transient imaging, and Velten et al.~\cite{Velten:2012:Visualizing} showed the first experimental NLOS imaging results using a femto-second laser and streak camera system. Since these seminal works, a growing body of work has been exploring transient imaging with a focus on enabling improved NLOS imaging~\cite{Pandharkar:2011,Naik:2011,Wu:2012,Gupta:12,heide2014diffuse,heide2013low,buttafava2015non,lindelltransient2017}.

\vspace{0.8em}\noindent\textbf{Impulse Non-Line-of-Sight-Imaging}
A major line of research~\cite{Pandharkar:2011,Velten:2012recovering,Gupta:12,o2018confocal,tsai2017geometry,arellano2017fast,pediredla2017reconstructing,OToole:2018,Xu:18} proposes to acquire transient images directly, by sending pulses of light into the scene and capturing the response with detectors capable of high temporal sampling. While the streak camera setup from Velten et al.~\cite{Velten:2012:Visualizing} allows for temporal precision of $< 10$~ps, corresponding to a path length of 3~mm, the high instrumentation cost and sensitivity has sparked work on single photon avalanche diodes (SPADs) as a detector alternative~\cite{buttafava2015non,OToole:2018}. Recently, O'Toole et al.~\cite{OToole:2018} propose scanned SPAD capture setup that allows for computational efficiency by modeling transport as a shift-invariant convolution. Although SPAD detectors can offer comparable resolution $<$ 10~ps~\cite{nolet2018quenching}, they typically suffer from low fill factors typically around a few percent~\cite{parmesan20149} and low spatial resolution in the kilo-pixel range~\cite{maruyama2011time}. Compared to ubiquitous intensity image sensors with $>$ 10 megapixel resolution, current SPAD sensors are still five orders of magnitude more costly, and two orders of magnitude less photon-efficient.

\vspace{0.8em}\noindent\textbf{Modulated and Coherent Non-Line-of-Sight-Imaging}
As an alternative to impulse-based acquisition, correlation time-of-flight setups have been proposed \cite{heide2013low,kadambi2013coded,heide2014diffuse,kadambi2016occluded} which encode travel-time indirectly in a sequence of phase measurements. While correlation time-of-flight cameras are readily available, e.g. Microsoft's Kinect One, their application to transient imaging is limited due to amplitude modulation bandwidths around 100~MHz, and hence temporal resolution in the nanosecond range. A further line of work~\cite{katz2012looking,katz2014non} explores using correlations in the carrier wave itself, instead of amplitude modulation. While this approach allows for single-shot NLOS captures, it is limited to scenes at microsopic scales~\cite{katz2014non}.

\vspace{0.8em}\noindent\textbf{Tracking and Classification}
Most similar to the proposed method are recent approaches that use conventional intensity measurements for NLOS vision tasks~\cite{klein2016tracking,caramazza2018neural,chan2017non,bouman2017turning}. Although not requiring temporal resolution, these existing approaches are restricted to coarse localization and classification to a limited extent, in contrast to full imaging and geometry reconstruction applications.

%% file: image_formation.tex
Non-line-of-sight imaging methods recover object properties outside the direct line-of-sight from third-order bounces. Typically, a diffuse wall patch in the direct line-of-sight is illuminated, where the light then scatters and partially reaches a hidden object outside the direct line-of-sight. At the object surface, the scattered light is reflected back to the visible wall where it may be measured. In contrast to existing methods that rely on temporally resolved transport, the proposed method uses stationary third-bounce transport, i.e. without time information, to recover reflectance and geometry of the hidden scene objects.
\subsection{Stationary Light Transport}
Specializing the Rendering Equation~\cite{Kajiya:1986:RE} to non-line-of-sight imaging, we model the radiance $\Radiance$ at a position $\wallpos$ on the wall as
\begin{equation}\label{eq:transport_model}
\begin{aligned}
	\Radiance(\wallpos)  = \hspace{-3pt} \int_{\Omega} & \,  \Refl \left( \voxelpos - \laserpos, \wallpos - \voxelpos \right) (\normal_\voxelpos \hspace{-3pt} \cdot \hspace{-1pt} (\voxelpos - \laserpos)) \frac{1}{r_{\voxelpos\wallpos}^2} \frac{1}{r^2_{\voxelpos\laserpos}} \Radiance(\laserpos) \de \voxelpos \\ &\hspace{-3pt}  + \delta\left(\|\laserpos-\wallpos\|\right) \Radiance(\laserpos) ,
\end{aligned}
\end{equation}
with $\voxelpos, \normal_\voxelpos$ the position and corresponding normal on the object surface $\Omega$, $\laserpos$ being a given beam position on the wall, and $\Refl$ denoting the bi-directional reflectance distribution function (BRDF). This image formation model assumes three indirect bounces, with the distance function $r$ modeling intensity falloff between input positions, and one direct bounce, when $\laserpos$ and $\wallpos$ are identical in the Dirac delta function $\delta( \cdot )$, and it ignores occlusions in the scene outside the line-of-sight. We model the BRDF with a diffuse and specular term as
\begin{equation}
\Refl \left( \omega_i, \omega_o \right) =  \alpha_d \, \Refl_d \left( \omega_i, \omega_o \right) \,\, + \,\, \alpha_s \, \Refl_s \left( \omega_i, \omega_o \right).
\end{equation}
The diffuse component $\Refl_d$ models light scattering, resulting in almost orientation-independent low-pass reflections without temporally coded illumination. In contrast, the specular reflectance component $\Refl_s$ contributes high-frequency specular highlights, i.e. mirror-reflections blurred by a specular lobe. These two components are mixed with a diffuse albedo $\alpha_d$ and specular albedo $\alpha_s$. While the spatial and color distributions of these two albedo components can vary, they are often correlated for objects composed of different materials, changing only at the boundaries of materials on the same surface. 
Although the proposed method is not restricted to a specific BRDF model, we adopt a Phong model~\cite{phong1975illumination} in the following.

\subsection{Sensor Model}
We use a conventional color camera in this work. We model the raw sensor readings with the Poisson-Gaussian noise model from Foi et al.~\cite{foi08,foi2009clipped} as samples
\begin{equation}\label{eq:imaging_PSF}
  b \, \sim \, \frac{1}{\kappa} \, \mathcal{P}\hspace{-2pt}\left( \text{$\int_{T} \int_{W} \int_{\Omega_A}$} \hspace{-5pt} \Radiance(\wallpos) \, \de \omega \, \de \wallpos \, \de t \,\, \frac{\kappa}{E} \right) \, + \, \mathcal{N}(0,\sigma^2), \\
\end{equation}
where we integrate Eq.~\eqref{eq:transport_model} over the solid angle $\Omega_A$ of the camera's aperture, over spatial position $W$ that the given pixel maps to, and exposure time $T$, resulting in the incident photons when divided by the photon energy $E$. The sensor measurement $b$ at the given pixel is then modeled with the parameters $\kappa > 0$ and $\sigma > 0$ in a Poisson and Gaussian distribution, respectively, accurately reflecting the effects of analog gain, quantum efficiency and readout noise. For notational brevity, we have not included sub-sampling on the color filter array of the sensor.

%% file: planar.tex
\begin{figure}[t!]
\vspace{-0.5cm}
  \centering
  \includegraphics[width=\linewidth]{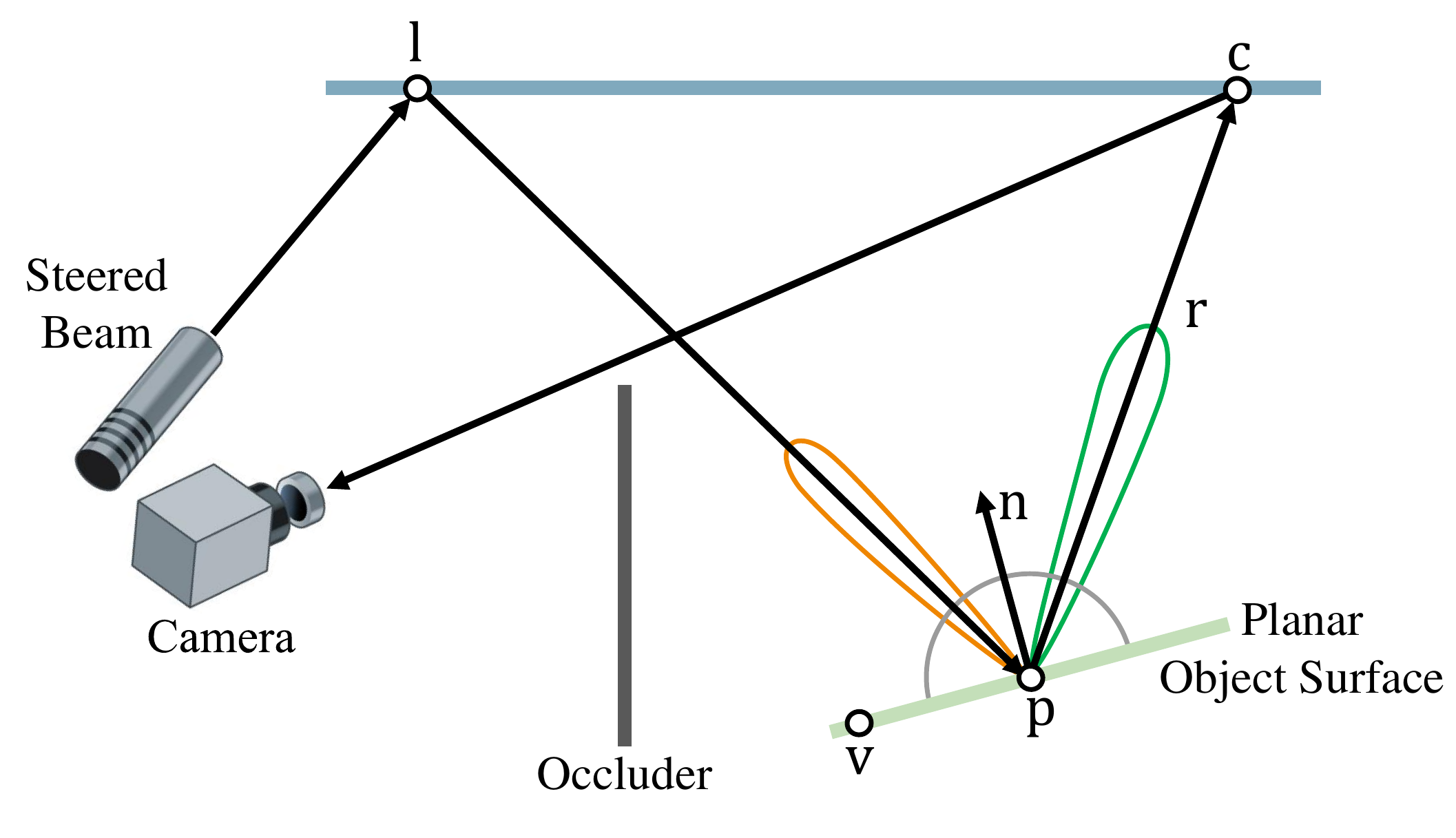}
  \vspace{-0.8cm}
  \caption{\small Indirect reflection on planar surface. A virtual source $\laserpos$ on the diffuse wall indirectly illuminates $\planepoint$ on a planar scene surface with $\normal$ and position $\planepos$. Depending on the surface BRDF, some light will be scattered back diffusely (uniformly in gray), retro-reflected back around $\laserpos$ (red lobe), and specularly reflected in direction $\Vect{r}$ to a visible point $\reflpoint$.}
   \label{fig:planar_brdf}
	\vspace{-0.3cm}
\end{figure}

In this section, we address the special case of planar objects. Assuming planar scenes in the hidden volume allows us to recover reflectance and 3D geometry from indirect reflections. Moreover, in this case, we can formulate the corresponding inverse problem using efficient optimization methods with analytic gradients. In the remainder of this paper, we assume that the shape and reflectance of the directly visible scene parts are known, i.e. the visible wall area. The proposed hardware setup allows for high-frequency spatially coded illumination, and hence the wall geometry can be estimated using established structured-light methods~\cite{scharstein2003high}.
Illuminating a patch $\laserpos$ on the visible wall, a hidden planar scene surface produces a diffuse low-frequency reflection component, encoding the projected position independently of the orientation~\cite{klein2016tracking}, and higher-frequency specular reflection components of the blurred specular albedo mapped to orientation-dependent positions on the wall. Assuming a single point light source at $\laserpos$ on the wall, see Fig. \ref{fig:planar_brdf}, the specular direction at a plane point $\planepoint$ is the mirror direction $\Vect{r} = (\planepoint - \laserpos) - 2( (\planepoint - \laserpos) \cdot \normal) \normal$ with the plane normal being $\normal$. The center of the specular lobe $\reflpoint$ on the wall is the mirror point of $\laserpos$, i.e. the intersection of the reflected ray in direction $\Vect{r}$ with the wall. Conversely, if we detect a specular lobe around $\reflpoint$ in a measurement, we can solve for the corresponding plane point as
\vspace{-3pt}
\begin{equation}\label{eq:corr_point}
\begin{aligned}
	\planepoint(\planepos, \normal)\hspace{-1pt} = \hspace{-1pt}\reflpoint + ( (\planepos\hspace{-1pt} - \hspace{-1pt}\reflpoint) \cdot \normal) \hspace{-1pt}\left(\normal\hspace{-1pt} - \hspace{-1pt}\frac{\planepos\hspace{-1pt} - \hspace{-1pt}\laserpos - ( (\reflpoint - \laserpos) \cdot \normal)\normal }{\normal \cdot (2\planepos - \reflpoint - \laserpos)} \right)\hspace{-2pt},
\end{aligned}
\end{equation}
that is a function of the planar surface represented by its normal $\normal$ and a point $\planepos$ on the plane. Eq.~\eqref{eq:corr_point} follows immediately from the constraint that the orthogonal projections of the points $\laserpos$ and $\reflpoint$ onto the plane result in equal triangles with $\planepoint$ and the respective point, see Supplemental Material for a detailed derivation. The plane has three degrees of freedom (DOF), which we parametrize as
\begin{equation}\label{eq:plane_parametrization}
	\vspace{-3pt}
	\normal(\theta,\phi) = \begin{bmatrix}\cos(\theta) \, \sin(\phi)\\ \sin(\theta) \, \sin(\phi)\\ \cos(\theta)\end{bmatrix}, \quad \planepos(\nu) = \Vec{o} + \nu\begin{bmatrix}0\\0\\1\end{bmatrix}.
	\vspace{-1pt}
\end{equation}
Expressing the normals in spherical coordinates ensures unit-length normals without explicit constraints. The plane position here is modeled as z-axis offset from the volume origin $\Vec{o}$.
With this parametrization we can estimate the specular albedo $\alpha_s$ and plane geometry $\theta, \phi, \nu$. Specifically, we sequentially illuminate the wall in $N$ spots uniformly sampled on the visible wall area and acquire a capture of the wall for each of the illumination points. Next, we extract sparse features and perform feature matching between the capture $\meas_t$ with the most detected features and all other captures $\meas_{\{1,\ldots,N\} \setminus {t} }$. We use SIFT features~\cite{lowe1999object} and RANSAC~\cite{fischler1981random} matching. Now for every feature $f \in \{1,\ldots,F\}$, this results in a set of matched positions $\reflpoint_i^f$ with $i \in \Psi_f$, and $\Psi_f$ being here the set of images with matches for feature $f$. We select the top $\tilde{F}$ features, with descending number of matches, and solve for the plane geometry by minimizing the reprojection distance on the plane as
\begin{equation}\label{eq:optimization}
\vspace{-3pt}
\begin{aligned}
\theta^*, \phi^*, \nu^*	 = \hspace{2pt} & \minimize{\theta, \phi, \nu} \hspace{-2pt} \sum_{f=1}^{\tilde{F}} \; \sum_{i \in \Psi_f} \left\| \planepoint^f_i(\planepos, \normal) - \overline{\planepoint^f} \right\|_2^2  \\
& \text{with} \quad \overline{\planepoint^f} = \frac{1}{|\Psi_f|}\sum_{i \in \Psi_f} \planepoint^f_i(\planepos, \normal),\\
\end{aligned}
\vspace{-1pt}
\end{equation}
where we use the notational shortcut $\planepoint^f_i(\planepos, \normal)$ for the plane point from Eq.~\ref{eq:corr_point} with reflected point $\reflpoint_i^f$ and laser point $l_i$. With this objective, we solve for consensus between the reprojected points on the plane for all features. The variable $\overline{\planepoint^f}$ represents here the mean position of all reprojected points for a given feature $f$. We solve the optimization problem from Eq.~\eqref{eq:optimization} using limited-memory BFGS~\cite{liu1989limited} which is a highly efficient quasi-Newton method. The analytic gradient of the least-squares objective from Eq.~\eqref{eq:optimization}, i.e. the partials w.r.t. $\theta, \phi, \nu$, are derived in the Supplemental Material.
Given the planar geometry estimate $\theta^*, \phi^*, \nu^*$, we estimate homography mappings $\Mat{H}_i$ between every image plane $i$ and the reprojected plane bounds. Specifically, we reproject the four image plane corners for measurement $\meas_i$ on the plane using Eq.~\ref{eq:corr_point} with laser position $\laserpos_i$ and reflection points as the four image plane corners. Given the homographies, we estimate the plane reflectance as the following maximum likelihood estimation problem
\begin{equation}\label{eq:analytic_solver}
\begin{aligned}
		\minimize{\Vect{x}} & - \sum_{i=1}^N \textrm{log} \left(\; p( \Vect{\meas_i} | \Mat{W}_{\Mat{H}_i} \convmatrix^\beta_{\Mat{H}_i} \Vect{x} ) \; \right) \; + \; \Gamma_{\text{TV}} \left( \Vect{x} \right) \\
		& \textrm{subject to} \quad 0 \leq \Vect{x}
\end{aligned}
\end{equation}
which is a linear inverse problem with $p( \Vect{\meas_i} | \Mat{W}_{\Mat{H}_i} \convmatrix^\beta_{\Mat{H}_i} \Vect{x})$ as the likelihood of observing a measurement $\meas_i$ given a specular reflectance $\Vect{x}$ on the plane, and $\Gamma_{\text{TV}} \left( \Vect{x} \right)$ as total variation (TV)~\cite{shin2016photon} prior on the specular reflectance itself. We follow~\cite{figueiredo2010restoration, lindelltransient2017} and assume a Poissonian-Gaussian likelihood term $p$ and adopt their variant of the Alternating Direction Method of Multipliers (ADMM)~\cite{boyd2011distributed} to solve the resulting linear inverse problem. While the solver method is established and described in detail in the Supplemental Material, our linear forward operator $\Mat{W}_{\Mat{H}_i} \convmatrix^\beta_{\Mat{H}_i} $ is the main difference in the proposed approach. This forward operator consists of the spatially varying convolution matrix  $\convmatrix$ which blurs the specular reflectance based on the angular falloff $\beta$ and distance to the wall (encoded by the homography), and a subsequent warping matrix $\Mat{W}$ which warps plane coordinates to the image plane and resamples the blurred specular reflectance coefficients using bi-linear interpolation. Although the proposed method is general, we assume a Gaussian spectral falloff with known standard deviation $\beta$. Note that the warping and falloff operators are only linear once the plane geometry is known.
\begin{figure*}[t!]
\vspace{-1.0cm}
    \centering
		\includegraphics[width=\textwidth]{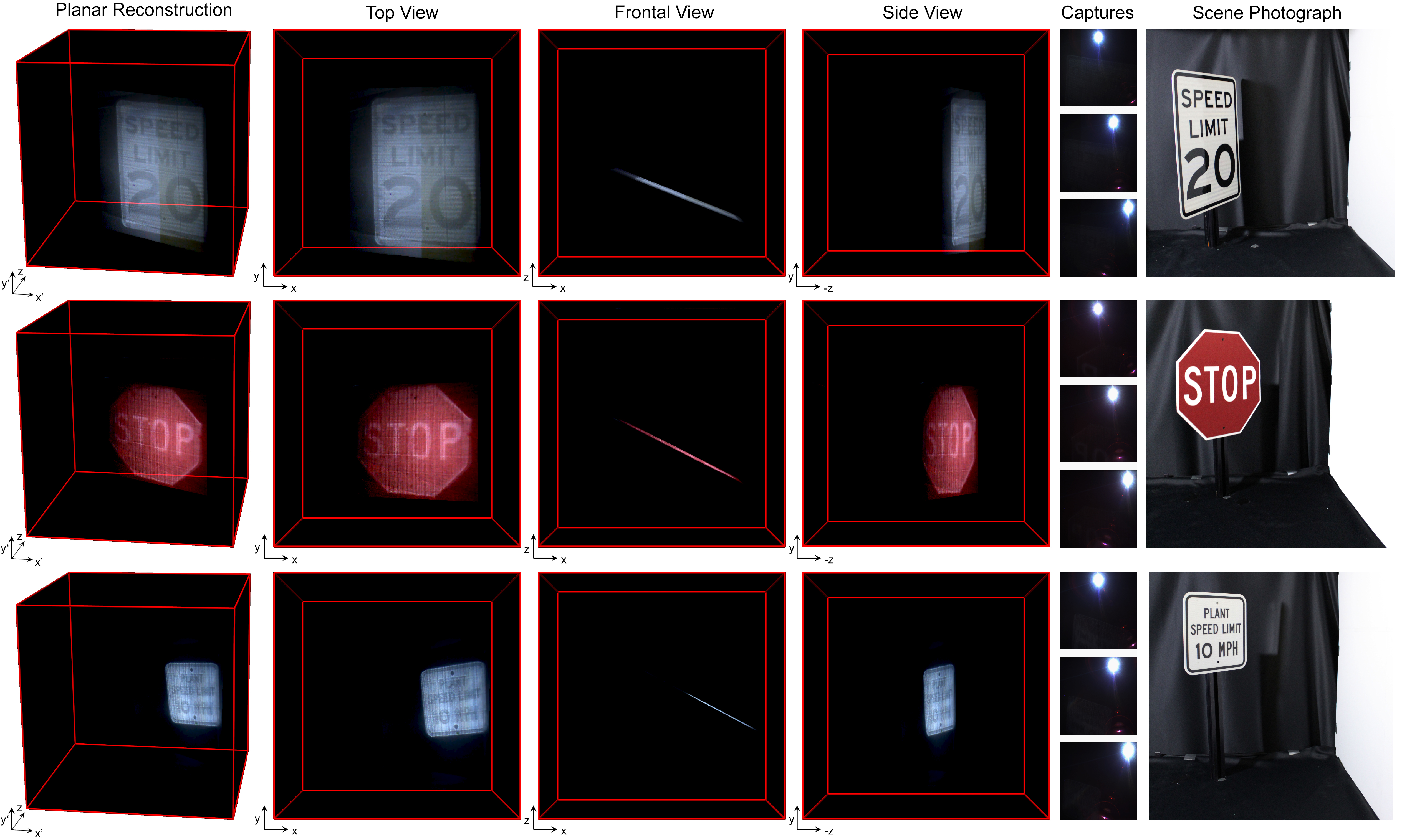}
    \vspace{-0.5cm}
    \caption{\small Experimental geometry and albedo reconstructions for the special case of planar objects, captured with the protoype from Sec.~\ref{sec:setup} and setup geometry from~\cite{OToole:2018}. We demonstrate reconstructions for three different surface materials. The first row shows an object with diamond grade retroreflective surface coating as they are found on number plates and high-quality street signs, identical to the objects in~\cite{OToole:2018}, which surprisingly contain faint specular components visible in the measurements (please zoom into the electronic version of this document). The second and third rows show a conventionally painted road sign and an engineering-grade street sign. The proposed method runs at around two seconds including capture and reconstruction, and achieves high resolution results without temporal sampling.}
    \label{fig:planar_reconstruction}
	\vspace{-0.3cm}
\end{figure*}

%% file: method.tex
In this section, we describe a trainable network and loss functions that allow us to recover occluded objects with arbitrary shape from intensity images. In contrast to planar geometries with only three DOF, arbitrary objects can have complex shapes with orders of magnitudes more parameters, i.e. for non-parametric surfaces three DOF per surface patch. To tackle this reconstruction challenge, we rely on strong priors on scene semantics which recent deep neural networks have been shown to encode efficiently~\cite{badrinarayanan2015segnet,ronneberger2015u}. 

%
%

%
\vspace{-8pt}
\paragraph{Input and Latent Parametrization} In the proposed capture setup, we project individual beams of light on the diffuse wall and capture the intensity image that is formed by steady-state global illumination in the scene. In general, projecting light beams to different positions on the wall results in different observations which we dub indirect reflection maps, i.e. indirect component of the image on the wall without the direct reflection. Each map contains information about the object shape and normal information in specific direction if the BRDF is angle-dependent. Note that this is not only the case for highly specular BRDFs, but also for lambertian BRDFs due to foreshortening and varying albedo. Hence, by changing the beam position we acquire variational information about shape and reflectance.

Assuming locally smooth object surfaces, we sample the available wall area uniformly in a 5~$\times$~5 grid and acquire multiple indirect reflection maps. We stack all the captured images, forming a $h \times w \times (5 \cdot 5 \cdot 3)$ dimension tensor as network input. The virtual source position is a further important information that may be provided to the network. However, since we use uniform deterministic sampling, we found that the model learns this structured information, in contrast to random source sampling.

We use the orthogonal view of the scene as our ground truth latent variable, as if the camera had been placed in the center of the visible wall in wall normal direction and with ambient illumination present. Given the stack of indirect reflection maps, the proposed network is trained to estimate the corresponding orthogonal view into the hidden scene.

\begin{figure*}[t!]
    \vspace{-1.0cm}
    \centering
    \includegraphics[width=0.95\textwidth]{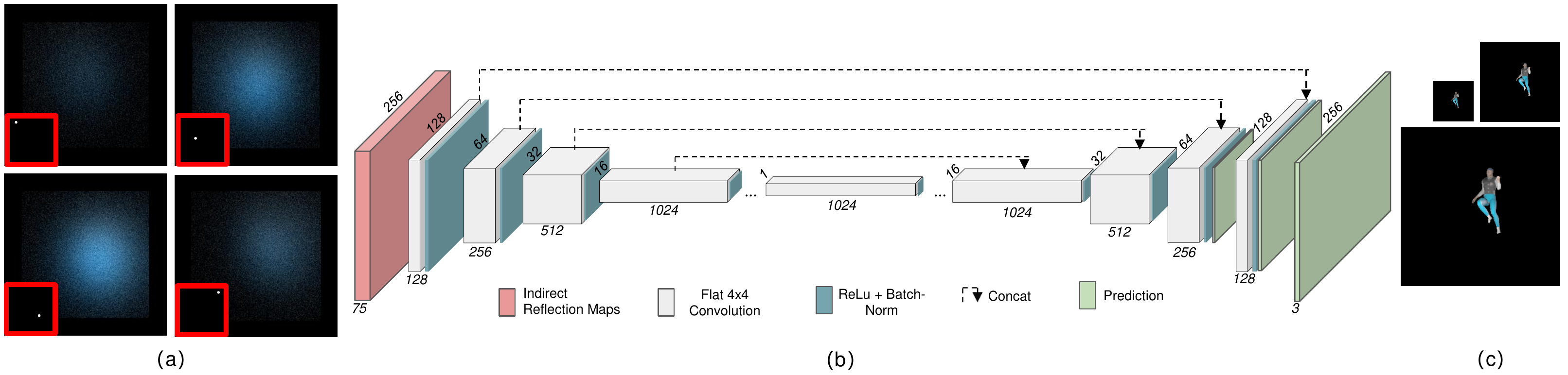}
    \vspace{-0.3cm}
    \caption{\small Learning NLOS imaging for arbitrary scene geometries and reflectance. We propose an encoder-decoder architecture that takes as input a stack of $5~\times~5$ synthetic indirect reflection measurements uniformly sampled on the diffuse wall. Here we show 4 examples of these 25 indirect reflection maps. The inset indicates the projected light beam position. The network outputs an orthogonal projection of the unknown scene as latent parametrization (c). We use a variant of the U-Net~\cite{ronneberger2015u} architecture (b) which predicts these projections at three scales.}
    \label{fig:Networks}
    \vspace{-0.3cm}
\end{figure*}

\vspace{-8pt}
\paragraph{Network Architecture} We propose a variant of the U-Net architecture~\cite{ronneberger2015u} as our network backbone structure, shown in Fig.~\ref{fig:Networks}. It contains a 8 layers encoder and decoder. Each encoder layer reduces the image size by a factor of two in each dimension and doubles the feature channel. This scaling is repeated until we retrieve a 1024 dimension latent vector. In corresponding convolution and deconvolution layer pairs with the same size, we concatenate them to learn residual information.

\paragraph{Loss functions}
We use a multi-scale $\ell_2$ loss function
\begin{align}
V_{multi-scale} & = \sum_{k} \gamma_k \| \Vect{i}^k - \Vect{o}^k \|^2, \label{eq:msmatch}
\end{align}
where $\Vect{i}$ is the predicted network output and $\Vect{o}$ is the ground-truth orthogonal image. Here, $k$ represents different scales and $\gamma_k$ is the corresponding weight of that layer. With feature map at $k$-the layer, we adopt an extra one deconvolution layer to convert the feature to an estimate at the target resolution. We predict 64~$\times$~64, 128~$\times$~128 and 256~$\times$~256 ground truth images and set the weights $\gamma_k$ as 0.6, 0.8 and 1.0. See the Supplemental Material for training details.

%% file: dataset.tex
The proposed deep model requires a large training set to represent objects with arbitrary shapes, orientations, locations and reflectance. While localization tasks for known objects~\cite{caramazza2018neural,chan2017non} may only require small datasets, that could be acquired experimentally, handling unknown scenes requires sampling a large combinatorial space, which we tackle by synthetic data generation. Although more practical than experimental acquisition, ray-traced rendering of indirect reflections still requires minutes per scene~\cite{jarabo14framework,Mitsuba}, and, for 5~min per measurement~\cite{Mitsuba}, a training set of 100,000~images would require one year of render time.
\begin{figure}[t]
\vspace{-2pt}
\hspace{-0.5cm}
  \centering
  \includegraphics[width=0.49\textwidth]{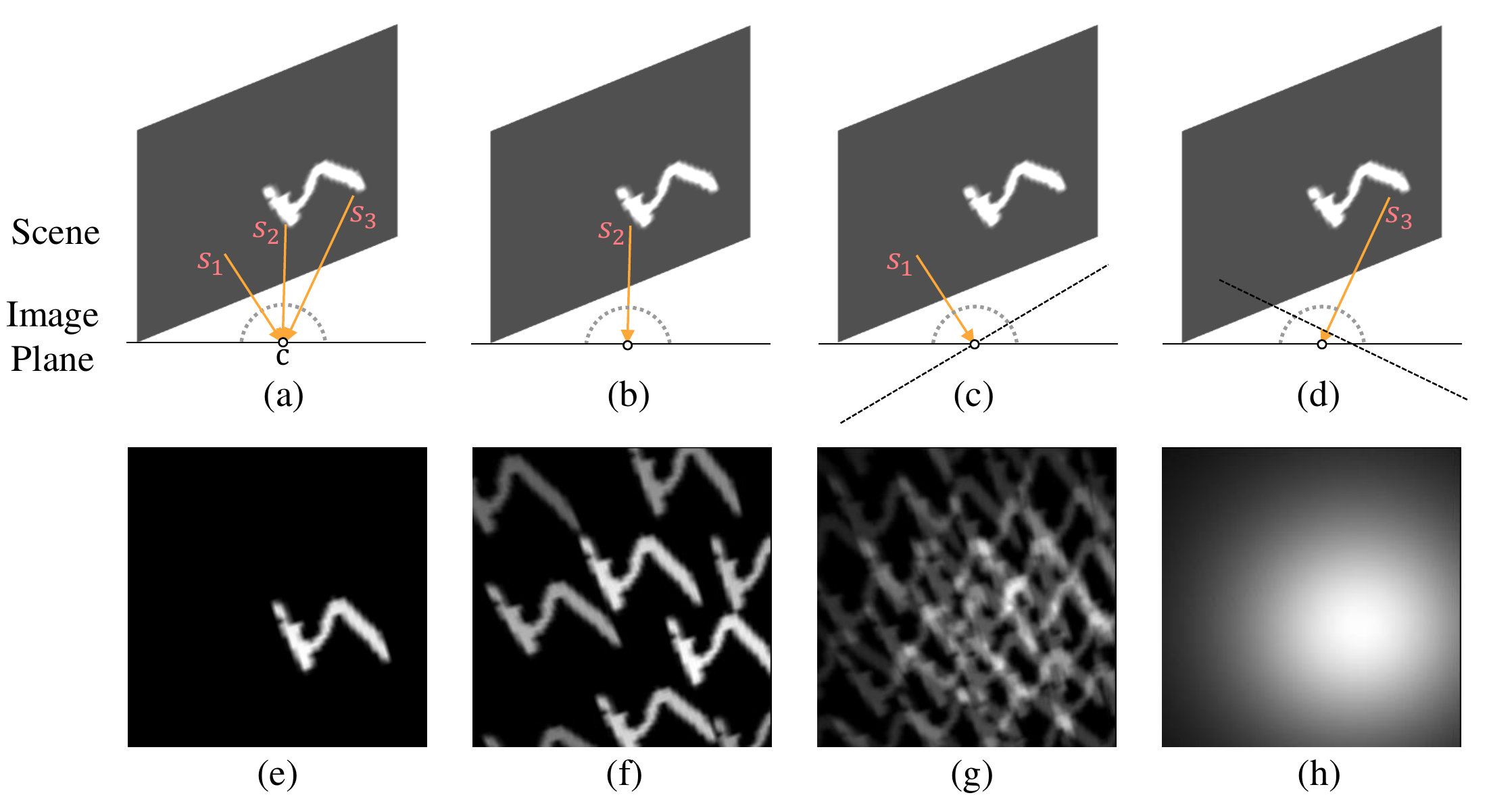}
	\vspace{-6pt}
  \caption{\small Training set rendering. Top row: (a) The radiance at a pixel $\reflpoint$ is the integral over the hemisphere of incoming indirect radiance, see Eq.~\eqref{eq:transport_model}. We uniformly sample the unit hemisphere (b-d), render each sample direction $s_i$ using direct rendering with orthographical projection, and accumulate the sample views. Bottom row: (e) Diffuse character with albedo sampled from MNIST. (f), (g) and (h) 5~$\times$~5, 10~$\times$~10, and 100~$\times$~100 directional samples.}
   \label{results:render_pipeline}
	\vspace{-10pt}
\end{figure}

Instead, we propose a novel pipeline for indirect third-bounce simulations using direct rendering. As shown in Fig.~\ref{results:render_pipeline}, a given wall pixel $\reflpoint$ integrates light over the hemisphere of incoming light directions, see Eq.~\eqref{eq:transport_model}. We sample the unit hemisphere to estimate this integral, rendering each sample direction using direct rendering with orthographical projection, and finally accumulating the sampled views. Hardware-accelerated OpenGL allows for microsecond render times for a single view direction. We synthesize a full third-bounce measurement in 0.1 seconds for 10000 hemisphere samples, which is more than 600$\times$ faster than~\cite{Mitsuba}, see Supplement Material. We render the training data adopting the setup geometry from~\cite{OToole:2018}. Fig.~\ref{results:rendered_images} shows examples for hidden objects from the following classes.

\vspace{0.7em}\noindent\textbf{MNIST Digits}
A number of recent works capture character-shaped validation objects~\cite{Velten:2012recovering,heide2013low,buttafava2015non,OToole:2018}. We represent this object class by placing MNIST digits on a plane, with randomly sampled rotations and shifts. We also sample specular coefficients in $[0,512]$ with a Phong~\cite{phong1975illumination} BRDF to represent different materials. We generate 20000 examples with albedo randomly sampled from MNIST.

\vspace{0.7em}\noindent\textbf{ShapeNet Models}
We synthesize measurements from ShapeNet~\cite{wu20153d} to represent more complex objects. 
We select the representative classes `car', `pillow', `laptop' and `chair', and train models for each class. Each class contains hundreds of models, and we render 20000 examples with random location, orientation, and reflectance as above.

\vspace{0.7em}\noindent\textbf{Human Models}
Finally, we synthesize data for human models with non-rigid, varying body shapes and poses. We sample these models from the SCAPE~\cite{anguelov2005scape} dataset and implement realistic appearance variation using the clothing samples from~\cite{Deep3DPose}. We generate 18000 examples with location-orientation sampling as above.

%% file: result.tex
In this section, we validate the proposed methods in simulation and using experimental measurements.
\subsection{Analysis}\label{synre}
\begin{figure}[t]
\vspace{-0.4cm}
  \centering
  \includegraphics[width=0.49\textwidth]{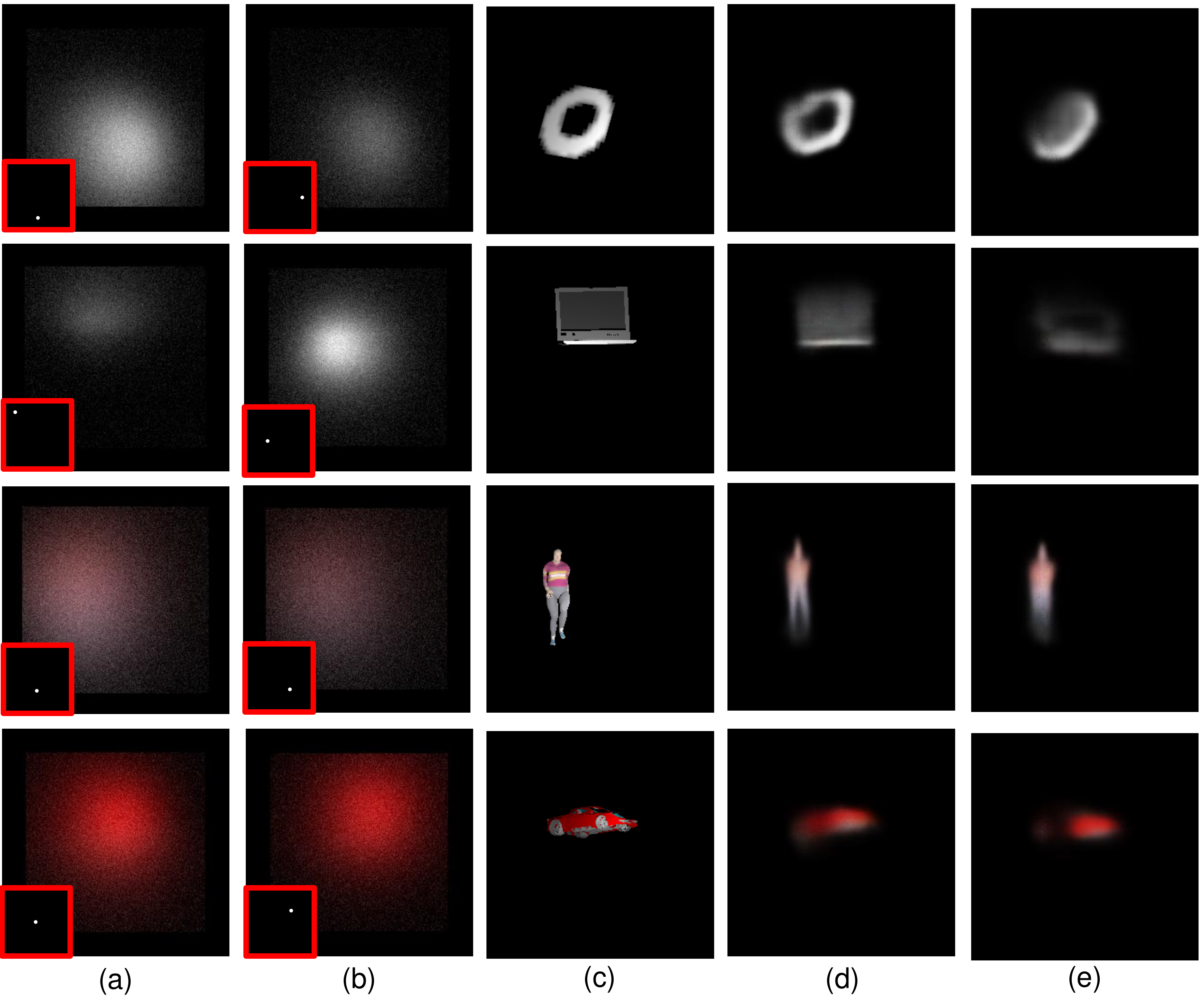}
	\vspace{-0.5cm}
  \caption{\small Qualitative results on synthetic data (including sensor noise simulation). (a) and (b) Two examples of 25 rendered indirect reflection maps. (c) Unknown scene (orthogonal projection). (d) and (e) Reconstruction by 5~$\times$~5 and 1~$\times$~1 indirect reflection maps, respectively.}
   \label{results:rendered_images}
	\vspace{-0.3cm}
\end{figure}
\begin{table}[t]
\vspace{0.1cm}
    \footnotesize
    \rowcolors{2}{white}{rowblue}
    \setlength{\tabcolsep}{3pt} 
    \setlength\extrarowheight{1pt}
    \centering
    \resizebox{\linewidth}{!}{

\begin{tabular}{ccccccc}
Beam Samples \hspace{-3pt} & MNIST & Car & Pillow & Laptop & Chair & Human \vspace*{2pt} \\ \vspace*{2pt}
 5~$\times$~5 & 24.0 & 26.3 & 26.7 & 25.9 & 22.5 & 25.5 \\ \vspace*{1pt}
 3~$\times$~3 & 22.5 & 25.9 & 25.9 & 24.8 & 22.3 & 25.3 \\ \vspace*{1pt}
 1~$\times$~1 & 21.4 & 25.9 & 24.8 & 24.6 & 22.1 & 25.0 \\
\end{tabular}
    }
		\vspace{-0.1cm}
    \caption{\small Reconstruction performance PSNR [dB] for decreasing virtual source sampling and varying object class. While the reconstruction performance drops significantly when reducing the source sampling, it does not completely fail even for a single source position. In this case, the method does not provide accurate shape, but only rough location information, see Fig.~\ref{results:rendered_images}. }
    \label{tab:re}
\vspace{-0.3cm}
\end{table}


We first assess the proposed network model, which runs at 32~FPS reconstruction rates, on unseen synthetic data from MNIST digits, ShapeNet, and the Human dataset. Fig.~\ref{results:rendered_images} shows two examples of indirect reflection maps and their corresponding light positions. We simulate Poissian-Gaussian noise $\sigma=0.05$ and $\kappa = 1/0.03$ according to Eq.~\eqref{eq:imaging_PSF}, calibrated for our experimental setup. The qualitative results in Fig.~\ref{results:rendered_images} show that the proposed model can precisely localize objects and recover accurate reflectance for large objects. Although recovery for smaller objects with diffuse reflectance becomes an extremely challenging task, our method still recovers coarse object geometry and reflectance. Table~\ref{tab:re} shows a quantitative evaluation of the reconstruction performance for different sampling patterns. While denser beam sampling on the wall results in higher recovery performance, even a single sampling position does provide enough information to perform recovery. However, Fig.~\ref{results:rendered_images} shows that fine geometry is lost in this case and only rough location and shape can be recovered.
\begin{figure}[t!]
\vspace{-0.5cm}
  \centering
  \includegraphics[width=\linewidth]{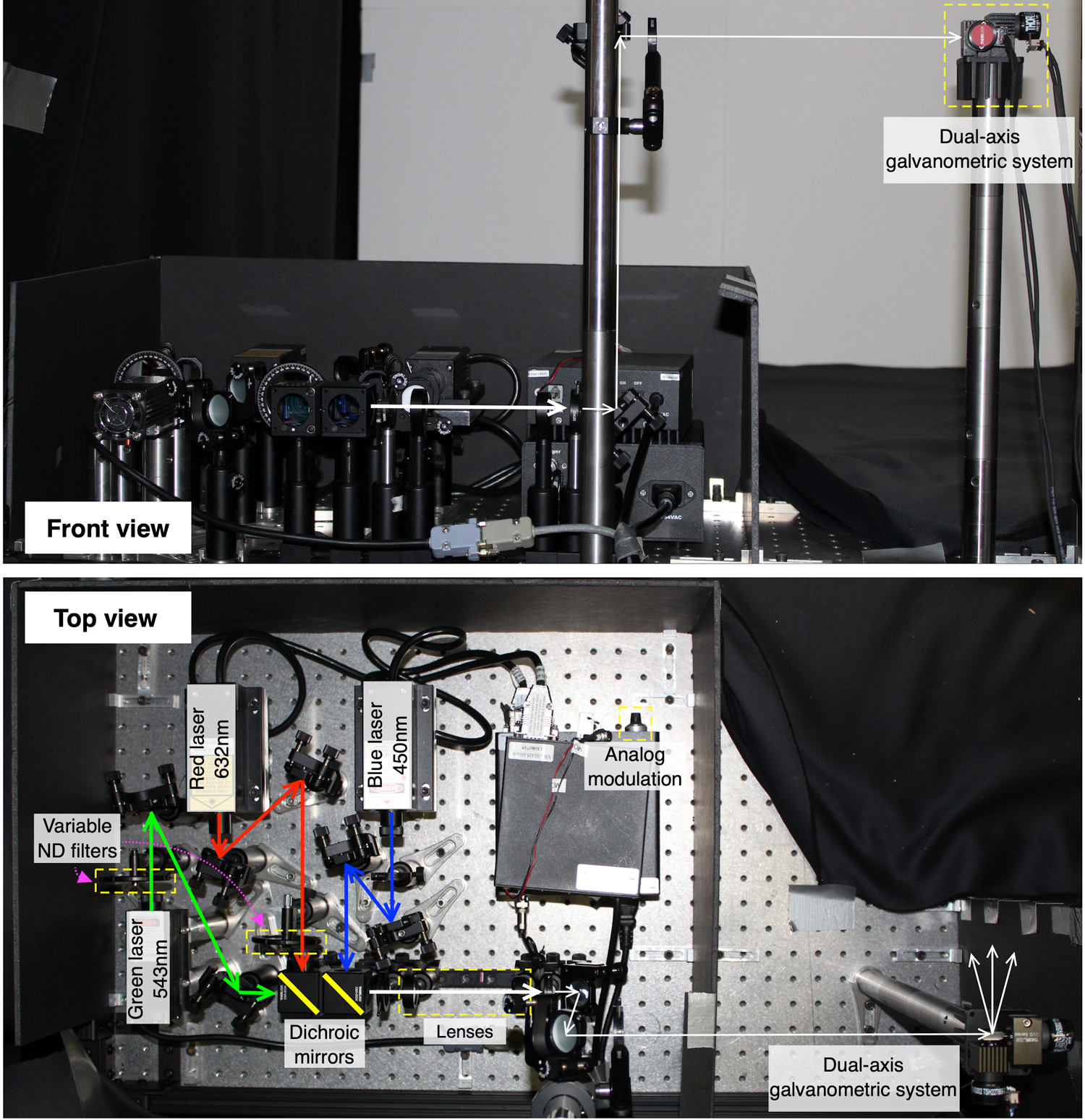}
  \caption{\small Prototype RGB laser source showing the path of the light, that ends in a dual-axis galvo. Top view: variable neutral density (ND) filters and one analog modulation is used to control the individual power of lasers. Long focal lenses combination and several irises reduce the beam diameter. Front view: only the white light path is shown. The camera is placed right next to the galvo.}
   \label{fig:Setup}
	\vspace{-0.3cm}
\end{figure}
\begin{figure*}[t!]
\vspace{-1.6cm}
  \centering
  \includegraphics[width=\textwidth]{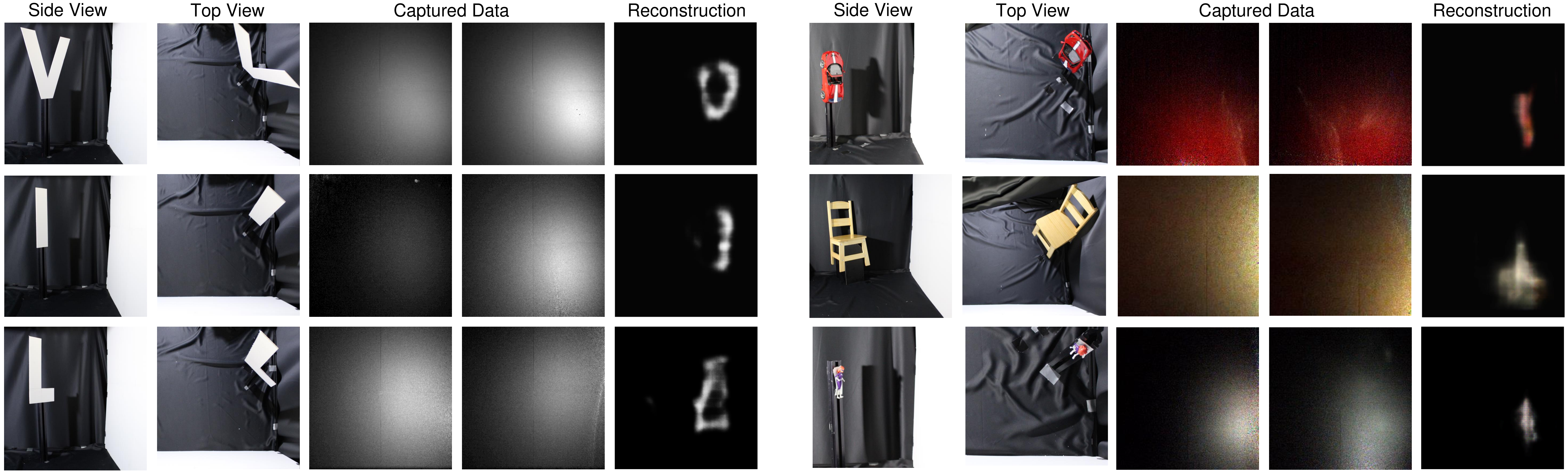}
  \caption{\small Non-line-of-sight reconstructions (orthogonal projections) from the proposed learned model for the setup geometry from~\cite{OToole:2018}. We demonstrate that the model can recover projected location on the wall and reflectance of fully diffuse scene objects such as the three diffuse letters on the left side. The scene objects are cutouts from diffuse foam-board material. The measurements have been contrast-normalized and the light position has been cropped to visualize the very low-signal diffuse component. The reconstructions on the right show captured scenes of more significantly smaller, more complex objects. For these challenging cases, the proposed method recovers rough shape, reflectance and projected position.}
   \label{results:realre}
	\vspace{-0.3cm}
\end{figure*}

\subsection{Experimental Setup}~\label{sec:setup}
To illuminate a hidden scene, we require a high-power, steerable and narrow white light source shown in Fig.~\ref{fig:Setup}. Unfortunately, these goals are challenging to achieve with wide sources like diodes (following from Gaussian beam considerations). We built a high-intensity collimated RGB light beam with three continuous 200~mW lasers of different wavelengths collinearized with two dichroic mirrors. Inexpensive modules with similar specifications are also available in a very small footprint as so-called RGB lasers. The wavelengths are chosen to correspond roughly to the maximums of the three-color sensor of our high-quantum-efficiency Allied Vision Prosilica GT 1930C camera (outside to the right of the Fig.). White balance is achieved by adjusting the power of each laser. Through the combination of several irises and lenses, the diameter of the beam is reduced to less than 5~mm, which is fed into a Thorlabs GVS102 dual-axis galvo. 
For each laser spot we acquire a single 50~ms exposure, leading to a full capture time of around 1.25~s for 5~$\times$~5 spots. Please see the Supplemental Material for additional details.

\subsection{Experimental Validation.}
Fig.~\ref{fig:planar_reconstruction} shows three planar reconstruction examples acquired using the described experimental setup. Surprisingly, even high-grade retroreflective surface coatings as they are found on number plates and high-quality street signs (also identical to the objects in~\cite{OToole:2018}), contain faint specular components visible in the measurements. The proposed optimization method achieves high quality reflectance and geometry recovery at almost interactive reconstruction times of about one second. While the dominating retroreflective component is returned to the virtual source position, a residual diffuse component is still present and appears as a halo-like artefact in the reconstruction. This diffuse halo is more apparent for the painted road sign, but still does not prohibit us from recovering high-quality geometry and reflectance for these planar objects without temporal coding.

Fig.~\ref{results:realre} shows reconstruction results for diffuse objects without requiring planar scene geometry. We demonstrate that the learned network, trained entirely on synthetic data, generalizes to experimental captures. The character-shaped objects on the left are cutouts of diffuse white foam boards, comparable to the validation scenes used in transient method~\cite{heide2013low,buttafava2015non,OToole:2018}. The proposed data-driven method accurately recovers shape and location on the wall for these large characters from their diffuse indirect reflections. For the smaller complex objects on the right, the diffuse reflections are substantially dimmer, but the proposed approach still recovers rough shape and position. Note that the mannequin figurine is only 3~cm in width and still recovered by the learned reconstruction method.

\begin{figure}[t]
    \vspace{-0.8em}
    \centering
    \includegraphics[width=\linewidth]{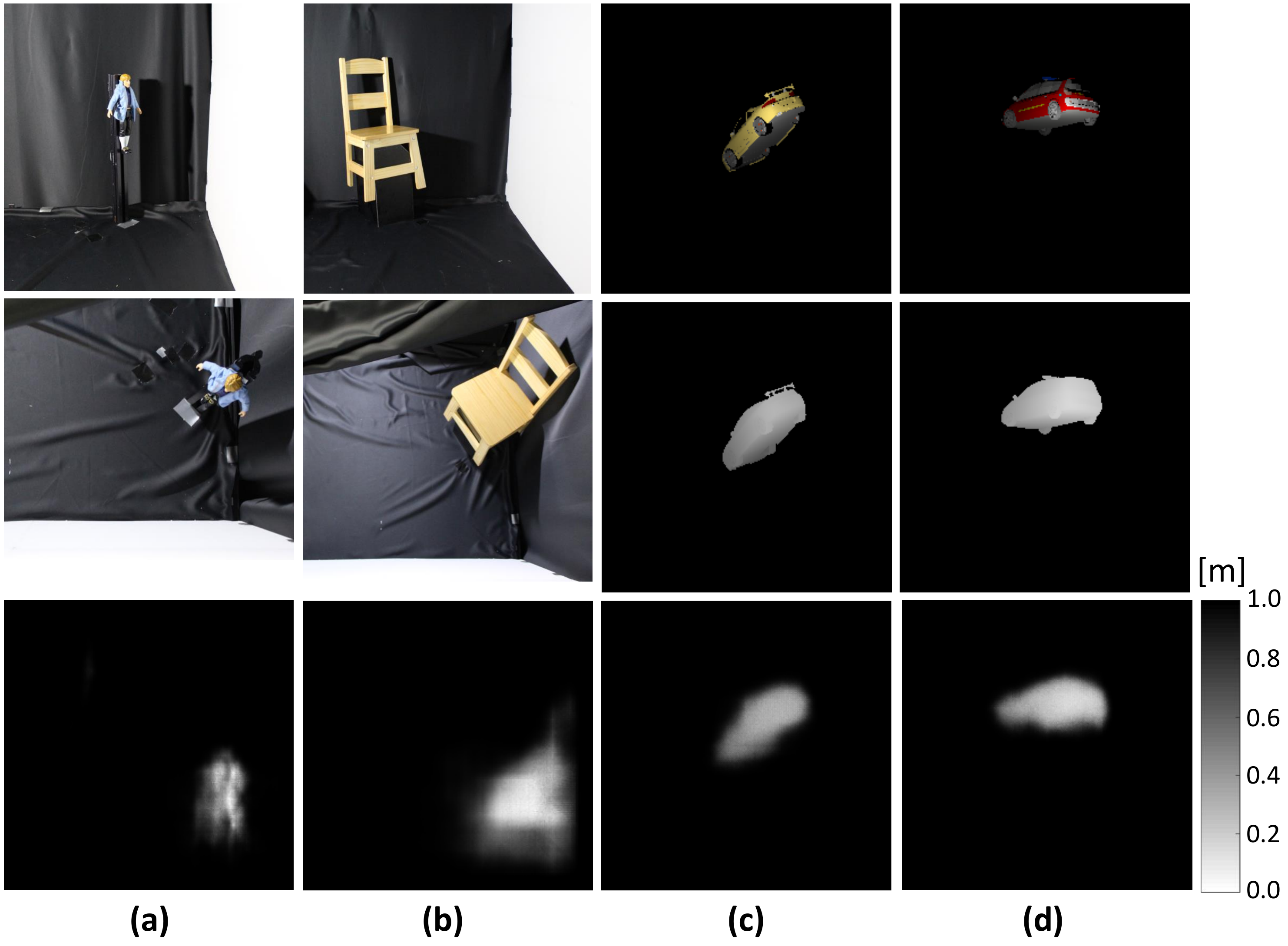}
    \vspace{-1.2em}
    \caption{\small Learning Hidden Geometry Reconstruction. The proposed method allows not only for albedo recovery, but also geometry reconstruction from steady-state indirect reflections. In (a,b), we show depth recovery of hidden scenes. From top to bottom: side and top view photographs of the hidden scene, and reconstructed depth map in meters. In (c,d) we show simulated results. From top to bottom: orthogonal view of hidden scene, ground truth depth of hidden scene, and reconstructed depth. }
    \vspace{-0.8em}
    \label{fig:depth}
\end{figure}

Finally, we demonstrate that the proposed method can also be applied to depth recovery of non-planar occluded scenes. Using the same network architecture as before, but with depth maps now as labels, the resulting model can recover reasonable object depth, which we validate in Fig.~\ref{fig:depth} for both synthetic and experimental measurements.

%% file: discussion.tex
We have demonstrated that it is possible to image objects outside of the direct line-of-sight using conventional RGB cameras and continuous illumination, without temporal coding. Relying on spatial variations in indirect reflections, we show high-quality NLOS geometry and reflectance recovery for planar scenes and a learned architecture which handles the shape-dependence of these variations. We have validated the proposed steady-state NLOS imaging method in simulation and experimentally. 

Promising directions for future research include the reduction of the laser spot size using single-model laser systems and achromatic optics, and mechanical occlusion before the lens system to discard the strong direct component along with lens flare. A further exciting opportunity are multiple inter-reflections in the hidden scene which may allow to conceptually turn every scene surface into a sensor. Relying on consumer color image sensors in our prototype system, the proposed method makes a first step towards this vision by achieving full-color non-line-of-sight imaging at fast imaging rates and in scenarios identical to those targeted by recent pulsed systems with picosecond resolution.